\documentclass[10pt,twocolumn]{article}

\usepackage[letterpaper,top=0.6in,bottom=0.6in,left=0.65in,right=0.65in,columnsep=0.2in]{geometry}
\usepackage{times}
\usepackage{latexsym}
\usepackage{booktabs}
\usepackage{multirow}
\usepackage{xcolor}
\usepackage{hyperref}
\usepackage{amsmath}
\usepackage{microtype}
\usepackage[numbers,square]{natbib}
\usepackage{graphicx}
\usepackage[compact]{titlesec}
\usepackage{float}

\definecolor{darkgreen}{rgb}{0,0.5,0}
\newcommand{\up}[1]{\textcolor{darkgreen}{+#1}}

\title{When Informal Text Breaks NLI:\\
Tokenization Failure, Distribution Shift, and Targeted Mitigations}

\author{Avinash Goutham Aluguvelly \\
  University of Texas at Austin \\
  Austin, TX, USA}

\date{}

\begin{document}
\maketitle

\begin{abstract}
We study how informal surface forms degrade NLI accuracy in ELECTRA-small
(14M) and RoBERTa-large (355M) across four transforms applied to SNLI
and MultiNLI: slang substitution, emoji replacement, Gen-Z filler tokens,
and their combination. Slang substitution (replacing formal words with informal equivalents,
e.g., \emph{going to} to \emph{gonna}, \emph{friend} to \emph{homie})
causes minimal degradation (at most 1.1pp):
slang vocabulary falls largely within WordPiece coverage, so the tokenizer
handles it without signal loss. Emoji replaces content words with Unicode
characters that ELECTRA's WordPiece tokenizer maps to \texttt{[UNK]},
destroying the input signal before any learned parameters see it (93.6\%
of emoji examples contain at least one \texttt{[UNK]}, mean 2.91 per
example). Noise tokens (\emph{no cap, deadass, tbh}) are fully
in-vocabulary but absent from NLI training data, consistent with the model assigning them inferential weight they do not carry. The two failure modes respond to different interventions:
preprocessing recovers emoji accuracy by normalizing text before
tokenization; augmentation handles noise by exposing the model to
noise-bearing examples during training. A hybrid of both achieves 88.93\%
on the combined variant for ELECTRA on SNLI (up from 75.88\%), with no
statistically significant drop on clean text. Against GPT-4o-mini zero-shot,
unmitigated ELECTRA is significantly worse on transformed variants
($p < 0.0001$); hybrid ELECTRA surpasses it across all SNLI variants and
reaches statistical parity on MultiNLI.
\end{abstract}

\section{Introduction}

NLI models trained on SNLI~\cite{bowman2015snli} and
MultiNLI~\cite{williams2018multinli} reach over 89\% and 90\% accuracy
respectively, yet these benchmarks consist almost entirely of formal,
carefully edited text. A premise in SNLI is an image caption; a hypothesis
is written by a crowdworker instructed to produce grammatical sentences.
Real text is messier. A user reviewing a product might write
\emph{``the jacket fits great no cap''} or send a message full of emoji
in place of the nouns a model was trained to reason over.

Prior robustness work in NLI has focused on label-level artifacts: cases
where models latch onto superficial correlations between words and labels
that arise from how datasets are constructed~\cite{gururangan2018annotation,
mccoy2019right}. We study a different vulnerability. The transforms we
apply preserve propositional content but change surface form in ways that
reflect how people actually write informally. If a model learned robust
semantic representations, accuracy should not change. When it does, that
reveals a dependence on specific surface patterns rather than meaning.

We apply three individual transforms (slang, emoji, noise) and their
combination across two models and two datasets. Slang barely affects
accuracy (at most 1.1pp drop). Emoji and noise each cause drops of up to
9pp on SNLI, reaching 13pp for the combined variant, with smaller but
consistent effects on MultiNLI (under 6pp). The more interesting finding
is why: emoji and noise fail through different mechanisms, which we
verify with direct tokenization analysis. That mechanistic difference
determines which mitigation works.

As a reference point, we compare against GPT-4o-mini zero-shot. Without
mitigation, fine-tuned ELECTRA is significantly worse on transformed
variants despite being stronger on clean text. Hybrid training closes
that gap.

\section{Related Work}

\subsection{Annotation Artifacts in NLI}

Several papers have shown that NLI models can exploit shallow cues in
standard benchmarks.~\citet{gururangan2018annotation} found that a model
trained only on hypotheses (no premises) achieves 67\% on SNLI by picking
up on lexical patterns introduced during annotation.~\citet{mccoy2019right}
showed that high lexical overlap between premise and hypothesis leads models
to predict entailment regardless of the actual relationship.~\citet{poliak2018hypothesis}
confirmed similar findings across multiple NLI datasets.

These are label-level artifacts: the problem is a spurious correlation
between surface features and labels. Our transforms are different.
We do not change labels or introduce new correlations; we change the
surface form while keeping the meaning constant. A model that genuinely
understands the semantics of a sentence should not care whether ``man'' is
spelled out or replaced by an emoji.

\subsection{Robustness to Surface Variation}

~\citet{gardner2020contrast} introduced contrast sets, minimal edits to
test examples that flip the gold label, finding accuracy drops of
10--25\%.~\citet{ribeiro2020checklist} proposed CheckList, a behavioral
testing framework that probes models with template-generated examples across
multiple linguistic phenomena.~\citet{morris2020textattack} developed
TextAttack for adversarial modifications to NLP inputs.

Our approach shares the spirit of these methods but differs in one key
respect: we do not change gold labels. The transforms we apply are
meaning-preserving, so any accuracy drop is entirely attributable to
surface form dependence rather than genuine semantic difficulty.

\subsection{Noisy and Informal Text}

~\citet{eisenstein2013phonological} documented systematic phonological
patterns in social media language.~\citet{baldwin2013noisy} studied
computational methods for processing noisy user-generated content. The
W-NUT shared tasks~\cite{derczynski2017results} benchmarked named entity
recognition on Twitter and other informal sources. Robustness to informal
text in NLI has received less attention than in sequence labeling tasks,
partly because the benchmark datasets are structurally formal. This paper
addresses that gap directly.

\subsection{Mitigation Approaches}

~\citet{liu2019inoculation} showed that fine-tuning on small amounts of
challenge set data can break artifact exploitation without requiring full
retraining.~\citet{zhou2020robustifying} demonstrated that diverse
augmentation improves generalization to out-of-distribution text. These
approaches focus on training-time interventions. We compare augmentation
against inference-time preprocessing and find that the two methods address
different failure modes, motivating a combined approach. Beyond testing
robustness, we identify distinct failure loci in the processing pipeline
and show that mitigation effectiveness depends on which stage fails.

\section{Methods}

\subsection{Datasets and Models}

We use \textbf{SNLI}~\cite{bowman2015snli} (550K train / 9,842 validation)
and \textbf{MultiNLI}~\cite{williams2018multinli} (393K train / 9,815
validation\_matched). SNLI pairs image-caption premises with crowdsourced hypotheses;
MultiNLI spans ten genres including telephone transcripts,
fiction, and government text, making it lexically more diverse.

We fine-tune two models. \textbf{ELECTRA-small}~\cite{clark2020electra}
has 14M parameters and uses a WordPiece tokenizer with a 30K vocabulary.
\textbf{RoBERTa-large}~\cite{liu2019roberta} has 355M parameters and uses
a BPE tokenizer with a 50K byte-level vocabulary. Both models are trained
for 3 epochs with batch size 32, learning rate $5 \times 10^{-5}$, and
max sequence length 128.

For RoBERTa augmented and hybrid runs, we found it necessary to add
\texttt{warmup\_ratio=0.06}. Without it, training collapsed to 33.82\%
accuracy (random guessing for a 3-class problem). ELECTRA-small did not
require warmup. We discuss this in Section~\ref{sec:stability}.

\subsection{Text Transforms}

All four transforms are applied to the evaluation set only. The training
data is clean; transform exposure comes through augmentation (described
below). Each transform preserves propositional content.

\noindent\textbf{Slang.} Phrase-level contractions (\emph{going to} $\to$
\emph{gonna}; \emph{trying to} $\to$ \emph{tryna}) and informal synonyms
(\emph{picture} $\to$ \emph{pic}; \emph{friend} $\to$ \emph{homie}).
Applied with probability 1.0. Covers roughly 35\% of SNLI vocabulary.

\noindent\textbf{Emoji.} Content words in 60+ noun and verb categories are replaced
by semantically related Unicode emoji (e.g., \emph{man} $\to$ \textit{[man
emoji]}; \emph{running} $\to$ \textit{[runner emoji]}). Multiple source words
can map to the same emoji, creating a many-to-one loss of information.
Covers roughly 40\% of SNLI vocabulary; on average 3.2 words are replaced
per example.

\noindent\textbf{Noise.} One Gen-Z affirmation or emphasis token is appended to the
hypothesis (\emph{deadass, lowkey, no cap, tbh, highkey, on god, frfr, real
talk, bet}). These tokens add no propositional content. None appear in
SNLI or MultiNLI training data. Applied with probability 1.0.

\noindent\textbf{Combined.} All three transforms applied simultaneously.

All transforms are designed to preserve propositional content. Slang and
emoji are applied to both premise and hypothesis; noise is appended to the
hypothesis only. Slang substitutions are meaning-equivalent by construction;
noise tokens add no propositional content. Emoji is the most ambiguous case:
a many-to-one mapping means some precision is lost, but the entailment
relationship holds as long as the emoji is semantically related to the
replaced word, which the mapping enforces.

Concrete before-and-after examples of each transform are shown in
Appendix~\ref{sec:transform-examples}.

\paragraph{Fixed evaluation sets.} We generate all five evaluation variants
(original plus four transforms) once, before any model training. All models
are evaluated on the exact same examples, which is required for valid paired
statistical tests.

\subsection{Mitigation Approaches}

\noindent\textbf{Augmentation.} Each training example is independently copied with
50\% probability and transformed using a randomly selected transform type.
This roughly doubles the training set (SNLI: 825K; MultiNLI: 590K). The
model sees the same semantic content in both formal and informal surface
forms, which we expect to build invariance to surface variation.

\noindent\textbf{Preprocessing.} Reverse transformations are applied at inference
time before the input reaches the tokenizer: slang is expanded to its
formal equivalent (98\% exact reconstruction), emoji are converted back to
their closest text label (73\% exact recovery due to many-to-one mapping),
and noise tokens are removed (100\% recall from our filter list). These
figures are computed by round-tripping each validation example through the
transform and its inverse. No retraining is required. The 73\% figure reflects an inherent many-to-one
limitation: a person emoji could have replaced \emph{man}, \emph{boy}, or
\emph{guy}, but preprocessing always maps it back to \emph{man}. The token
is no longer \texttt{[UNK]}, but it is not guaranteed to be the original
word.

\noindent\textbf{Hybrid.} Augmentation at training time and preprocessing at
inference time, combined. This requires both a retrained model and a
preprocessing layer, but the costs are low: preprocessing adds negligible
latency, and augmented training adds roughly 50\% more steps.

\subsection{LLM Baselines}

We evaluate \textbf{GPT-4o-mini} and \textbf{GPT-3.5-turbo} zero-shot on
the same fixed evaluation sets using a single-turn prompt:

\begin{quote}
\small\texttt{System: You are a natural language inference classifier.
Given a premise and hypothesis, output exactly one word: entailment,
neutral, or contradiction. No explanation.}
\end{quote}

Temperature is set to 0; responses are cached to ensure consistent evaluation outputs across analyses.

\subsection{Statistical Tests}

We use McNemar's test~\cite{mcnemar1947note} with continuity correction
for all pairwise model comparisons. Because all models are evaluated on
the same fixed examples, the test is valid. We apply Bonferroni correction
within two separate test families:

\begin{itemize}
\item Fine-tuned model comparisons: $\binom{4}{2} \times 5 = 30$ tests,
  threshold $\alpha = 0.05/30 \approx 0.0017$ (marked \textbf{**}).
\item LLM comparisons: treated as a separate family per model/dataset
  combination (5--10 tests each), threshold $\alpha = 0.005$--$0.01$.
  All key claims have $p < 0.0001$ and remain significant under the stated correction.
\end{itemize}

Bootstrap 95\% confidence intervals (2,000 replicates, seed 42) are
computed for all reported accuracy figures; margins are $\pm$0.3--0.5pp.

\section{Results}

\subsection{Baseline Performance}

Table~\ref{tab:baseline} shows accuracy under each transform with no
mitigation. Slang causes almost no degradation (at most 1.1pp for every
model and dataset). Emoji and noise are the main sources of failure, with
the combined variant producing the largest drops.

\begin{table*}[t]
\centering
\small
\begin{tabular}{lrrrrr}
\toprule
\textbf{Model / Dataset} & \textbf{Orig.} & \textbf{Slang} & \textbf{Emoji} & \textbf{Noise} & \textbf{Comb.} \\
\midrule
ELECTRA / SNLI    & 89.07 & 87.97 & 80.93 & 80.19 & 75.88 \\
ELECTRA / MNLI    & 81.51 & 80.83 & 80.97 & 75.54 & 76.11 \\
RoBERTa / SNLI    & 93.12 & 92.80 & 85.12 & 89.77 & 82.14 \\
RoBERTa / MNLI    & 90.55 & 90.06 & 89.80 & 87.57 & 87.99 \\
\midrule
GPT-4o-mini / SNLI & 87.90 & 87.33 & 83.01 & 86.26 & 81.59 \\
GPT-3.5 / SNLI     & 66.65 & 66.22 & 62.99 & 63.55 & 61.54 \\
GPT-4o-mini / MNLI & 82.55 & 82.55 & 82.36 & 81.85 & 81.53 \\
GPT-3.5 / MNLI     & 65.57 & 65.35 & 65.21 & 64.16 & 64.55 \\
\bottomrule
\end{tabular}
\caption{Accuracy (\%) with no mitigation applied. MNLI = MultiNLI.
  Bootstrap 95\% CI margins: $\pm$0.3--0.5pp across all cells.}
\label{tab:baseline}
\end{table*}

The gap between emoji and noise robustness differs substantially by model
scale. ELECTRA drops 8.1pp on emoji and 9.0pp on noise (SNLI). RoBERTa
drops a similar 8.0pp on emoji but only 3.4pp on noise. The emoji drop is
nearly identical across models; the noise gap is not. This asymmetry is
explained in Section~\ref{sec:mechanism}.

\subsection{Mitigation Results}

Tables~\ref{tab:electra-snli} through~\ref{tab:roberta} show results for
all four approaches. For ELECTRA, hybrid is best or tied-best across every
variant on both datasets. For RoBERTa on SNLI, the same pattern holds.
RoBERTa on MultiNLI is more mixed: preprocessing recovers emoji better than
hybrid does (90.31\% vs.\ 90.14\%), and augmentation beats hybrid on noise
(90.37\% vs.\ 90.17\%). The mitigation ordering mirrors the failure
mechanism: preprocessing helps emoji more, augmentation helps noise more.

\begin{table}[h]
\centering
\small
\begin{tabular}{lrrrrr}
\toprule
\textbf{Approach} & \textbf{Orig.} & \textbf{Slang} & \textbf{Emoji} & \textbf{Noise} & \textbf{Comb.} \\
\midrule
Baseline      & 89.07 & 87.97 & 80.93 & 80.19 & 75.88 \\
Augmentation  & 89.41 & 89.05 & 84.67 & \textbf{89.44} & 84.50 \\
Preprocessing & 89.05 & 88.93 & \textbf{88.71} & 82.36 & 84.43 \\
\textbf{Hybrid} & \textbf{89.24} & \textbf{89.13} & 88.96 & 89.26 & \textbf{88.93} \\
\bottomrule
\end{tabular}
\caption{ELECTRA-small on SNLI (\%). Bold marks the best result per column.
  Hybrid vs.\ baseline: $p < 0.0001^{**}$ for emoji, noise, and combined;
  $p > 0.05$ for original. Preprocessing beats augmentation on emoji
  ($p < 0.0001^{**}$); augmentation beats preprocessing on noise
  ($p < 0.0001^{**}$).}
\label{tab:electra-snli}
\end{table}

\begin{table}[h]
\centering
\small
\begin{tabular}{lrrrrr}
\toprule
\textbf{Approach} & \textbf{Orig.} & \textbf{Slang} & \textbf{Emoji} & \textbf{Noise} & \textbf{Comb.} \\
\midrule
Baseline      & 81.51 & 80.83 & 80.97 & 75.54 & 76.11 \\
Augmentation  & 81.78 & 81.50 & 81.48 & \textbf{81.61} & 81.00 \\
Preprocessing & 81.50 & 81.38 & \textbf{81.50} & 77.09 & 78.73 \\
\textbf{Hybrid} & \textbf{82.19} & \textbf{82.14} & 82.20 & 82.18 & \textbf{82.12} \\
\bottomrule
\end{tabular}
\caption{ELECTRA-small on MultiNLI (\%). Hybrid vs.\ baseline:
  $p < 0.0001^{**}$ for noise and combined.}
\label{tab:electra-mnli}
\end{table}

\begin{table*}[t]
\centering
\small
\begin{tabular}{llrrrrr}
\toprule
\textbf{Dataset} & \textbf{Approach} & \textbf{Orig.} & \textbf{Slang} & \textbf{Emoji} & \textbf{Noise} & \textbf{Comb.} \\
\midrule
\multirow{4}{*}{SNLI}
 & Baseline     & 93.12 & 92.80 & 85.12 & 89.77 & 82.14 \\
 & Augmentation & 92.94 & 92.91 & 92.74 & \textbf{92.91} & 92.78 \\
 & Preprocessing & 92.97 & 92.98 & \textbf{92.70} & 90.63 & 91.02 \\
 & \textbf{Hybrid} & \textbf{93.00} & \textbf{92.96} & 92.87 & 92.89 & \textbf{92.88} \\
\midrule
\multirow{4}{*}{MNLI}
 & Baseline     & 90.55 & 90.06 & 89.80 & 87.57 & 87.99 \\
 & Augmentation & 90.41 & 90.33 & 90.09 & \textbf{90.37} & 90.11 \\
 & Preprocessing & 90.33 & 90.26 & \textbf{90.31} & 87.78 & 88.97 \\
 & Hybrid       & 90.20 & 90.22 & 90.14 & 90.17 & \textbf{90.17} \\
\bottomrule
\end{tabular}
\caption{RoBERTa-large results (\%). On MultiNLI, individual mitigations
  outperform hybrid on specific variants, unlike the ELECTRA pattern.}
\label{tab:roberta}
\end{table*}

\subsection{Why Emoji and Noise Fail Differently}
\label{sec:mechanism}

The most informative signal in the mitigation results is not the top-line
numbers but the ordering: preprocessing beats augmentation on emoji, and
augmentation beats preprocessing on noise. This pattern holds across both
models and both datasets. It points to two different failure modes.

\paragraph{Emoji and tokenization.} We ran a tokenization analysis across
all five eval variants. For ELECTRA's WordPiece tokenizer, emoji characters
fall outside the 30K vocabulary entirely. They are mapped to
\texttt{[UNK]}. Table~\ref{tab:tokenization} shows that 93.6\% of
emoji-transformed examples contain at least one \texttt{[UNK]} token, with
a mean of 2.91 per example. Slang and noise produce zero \texttt{[UNK]}
tokens. The mechanism is direct: emoji destroys the input before the model
can process it. Preprocessing sidesteps this by converting emoji back to
text before tokenization; augmentation cannot recover information the
tokenizer has already discarded.

RoBERTa's BPE tokenizer covers all byte sequences and never produces
\texttt{[UNK]}; its \texttt{[UNK]} rate is 0\% for all variants. Yet
RoBERTa still drops 8.0pp on emoji. The fragmentation analysis explains
this: emoji inflates RoBERTa's subword-per-word ratio from 1.153 to 1.346
(+16.7\%), while noise increases it only +1.8\%. Emoji are broken into
multiple BPE tokens that carry no semantic signal, flooding the sequence
with noise of a different kind. Preprocessing still helps because removing
the emoji before tokenization prevents the fragmentation.

\begin{table*}[t]
\centering
\small
\begin{tabular}{lrrr}
\toprule
\textbf{Variant} & \textbf{ELECTRA UNK/ex} & \textbf{\% ex w/ UNK} & \textbf{RoBERTa subw/word} \\
\midrule
Original  & 0.000 &  0.0\% & 1.153 \\
Slang     & 0.000 &  0.0\% & 1.192 (\up{3.4\%}) \\
Emoji     & \textbf{2.909} & \textbf{93.6\%} & \textbf{1.346} (\up{16.7\%}) \\
Noise     & 0.000 &  0.0\% & 1.174 (\up{1.8\%}) \\
Combined  & 2.769 & 91.8\% & 1.371 (\up{18.9\%}) \\
\bottomrule
\end{tabular}
\caption{Tokenization statistics on SNLI ($n = 9{,}842$). ELECTRA
  WordPiece: mean \texttt{[UNK]} per example and percentage of examples
  containing at least one \texttt{[UNK]}. RoBERTa BPE: mean subword
  tokens per word (fragmentation rate).}
\label{tab:tokenization}
\end{table*}

\paragraph{Noise and distribution shift.} Noise tokens (\emph{deadass,
lowkey, no cap}) are in-vocabulary for both models. The zero \texttt{[UNK]}
rate (Table~\ref{tab:tokenization}) confirms this. The problem is not that
the tokenizer cannot represent them; it is that the model has no prior
experience with these tokens in an NLI context. SNLI and MultiNLI
training data contains only formal, edited text. When \emph{deadass}
appears in a hypothesis, the model tries to reason about it as a content
word, disrupting inference. Augmentation exposes the model to noise tokens
paired with their clean equivalents during training, teaching it to ignore
them. Preprocessing removes noise tokens mechanically, but a filter list
cannot generalize to novel tokens that were not anticipated.

The two failure modes produce distinct error signatures.
Table~\ref{tab:confusion} shows the predicted-label distribution for
each true class on the ELECTRA-small SNLI baseline. Emoji errors
collapse toward \emph{neutral}: contradiction examples are predicted
entailment (10\%) or neutral (12\%), the model loses the
contradiction signal when key tokens become \texttt{[UNK]}. Noise
errors push in the opposite direction: entailment examples are
predicted \emph{contradiction} at 14\%, and neutral examples at 17\%,
rates far above the 4\% and 9\% seen under emoji. Noise tokens
appended to the hypothesis appear to carry inferential weight,
pushing predictions toward contradiction. The error
direction is consistent with the mechanism: signal destruction
defaults to neutral; spurious tokens carry contradiction-like inferential weight.

\begin{table*}[t]
\centering
\small
\setlength{\tabcolsep}{6pt}
\begin{tabular}{l ccc ccc}
\toprule
 & \multicolumn{3}{c}{Emoji variant} & \multicolumn{3}{c}{Noise variant} \\
\cmidrule(lr){2-4}\cmidrule(lr){5-7}
True label & ENT & NEU & CON & ENT & NEU & CON \\
\midrule
Entailment    & \underline{85} & 11 &  4 & \underline{71} & 15 & \textbf{14} \\
Neutral       & 11 & \underline{80} &  9 &  6 & \underline{77} & \textbf{17} \\
Contradiction & \textbf{10} & \textbf{12} & \underline{78} &  2 &  5 & \underline{93} \\
\bottomrule
\end{tabular}
\caption{Predicted label distribution (\%) per true class, ELECTRA-small baseline on SNLI ($n = 9{,}842$).
  Diagonal (underlined) = correct predictions. Bold marks where the two transforms produce the most
  divergent errors: emoji pushes contradiction toward neutral and entailment; noise pushes entailment
  and neutral toward contradiction.}
\label{tab:confusion}
\end{table*}

\subsection{Scale Effects}

The comparison between ELECTRA and RoBERTa (Table~\ref{tab:baseline})
shows that scale helps with noise but not with emoji. RoBERTa's noise drop
on SNLI is 3.4pp against ELECTRA's 9.0pp; its emoji drop is 8.0pp against
ELECTRA's 8.1pp. The noise improvement is consistent with the hypothesis that RoBERTa's
larger pretraining corpus included more informal text, giving it some
implicit exposure to conversational tokens. Emoji is different: on SNLI, the drop is nearly identical across
models (ELECTRA 8.1pp, RoBERTa 8.0pp), and 25$\times$ more parameters
does not help. Both models also drop similarly on MultiNLI emoji,
though the absolute drops are smaller there (under 1pp for both).

The mitigation story also changes at scale. For ELECTRA, hybrid is
uniformly best. For RoBERTa on MultiNLI, preprocessing alone beats hybrid
on emoji and augmentation alone beats hybrid on noise. A larger model with
more pretraining vocabulary and corpus diversity benefits less from
combining mitigations; the targeted individual approach is often sufficient.

\subsection{Fine-Tuned Models vs.\ Zero-Shot LLMs}
\label{sec:llm}

Table~\ref{tab:llm} shows ELECTRA vs.\ GPT-4o-mini on SNLI with paired
significance tests. A fine-tuned ELECTRA-small without any mitigation is
significantly \emph{worse} than GPT-4o-mini on emoji, noise, and combined
($p < 0.0001$, Bonferroni-corrected). With hybrid training, all three
significant disadvantages reverse: hybrid ELECTRA beats GPT-4o-mini on
all five variants (Table~\ref{tab:llm}). On MultiNLI,
ELECTRA hybrid reaches statistical parity with GPT-4o-mini across all
variants (all $p > 0.24$).

\begin{table*}[t]
\centering
\small
\begin{tabular}{llrrrc}
\toprule
\textbf{Var.} & \textbf{ELECTRA} & \textbf{Acc.} & \textbf{GPT-4o-mini} & \textbf{$p$} & \\
\midrule
\multirow{2}{*}{Orig.}
 & Baseline & 89.07 & \multirow{2}{*}{87.90} & 0.0036 & FT$>$LLM$^{**}$ \\
 & Hybrid   & 89.24 &                         & 0.0008 & FT$>$LLM$^{**}$ \\
\midrule
\multirow{2}{*}{Slang}
 & Baseline & 87.97 & \multirow{2}{*}{87.33} & 0.1247 & NS \\
 & Hybrid   & 89.13 &                         & $<$0.0001 & FT$>$LLM$^{**}$ \\
\midrule
\multirow{2}{*}{Emoji}
 & Baseline & 80.93 & \multirow{2}{*}{83.01} & $<$0.0001 & FT$<$LLM$^{**}$ \\
 & Hybrid   & 88.96 &                         & $<$0.0001 & FT$>$LLM$^{**}$ \\
\midrule
\multirow{2}{*}{Noise}
 & Baseline & 80.19 & \multirow{2}{*}{86.26} & $<$0.0001 & FT$<$LLM$^{**}$ \\
 & Hybrid   & 89.26 &                         & $<$0.0001 & FT$>$LLM$^{**}$ \\
\midrule
\multirow{2}{*}{Comb.}
 & Baseline & 75.88 & \multirow{2}{*}{81.59} & $<$0.0001 & FT$<$LLM$^{**}$ \\
 & Hybrid   & 88.93 &                         & $<$0.0001 & FT$>$LLM$^{**}$ \\
\bottomrule
\end{tabular}
\caption{ELECTRA vs.\ GPT-4o-mini on SNLI (Bonferroni $\alpha = 0.005$,
  10 tests). $^{**}$ marks Bonferroni-significant comparisons.}
\label{tab:llm}
\end{table*}

RoBERTa-large outperforms GPT-4o-mini across nearly all conditions on both
datasets. The one exception is RoBERTa baseline on combined SNLI
(82.14\% vs.\ 81.59\%), where the difference is not significant;
all other RoBERTa comparisons are significant ($p < 0.0001$). Both fine-tuned models comfortably outperform
GPT-3.5-turbo on all conditions.

One additional pattern is worth noting. GPT-4o-mini drops 4.89pp on SNLI
emoji and 6.31pp on combined SNLI, but only 0.18pp and 1.02pp on the same
conditions in MultiNLI. SNLI's premise-hypothesis pairs come from a narrow
domain (image captions), so an emoji replacing a single content word can be
high-impact. MultiNLI's genre diversity spans telephone conversations,
fiction, and government documents, and GPT-4o-mini has likely seen more
varied text during pretraining. This suggests the robustness advantage of
frontier LLMs is partly domain-dependent rather than fundamental.

\subsection{Error Analysis}

We examined all cases where the baseline fails but the hybrid succeeds.
The most common pattern is what we call label collapse: when the input
contains \texttt{[UNK]} tokens (emoji variant) or unfamiliar tokens (noise
variant), the model defaults to predicting \texttt{neutral}. In the SNLI
emoji variant, the baseline produces entailment$\to$neutral errors 360 times
and contradiction$\to$neutral errors 405 times. The hybrid recovers 1,027
emoji cases and 1,117 noise cases, with improvement-to-regression ratios of
4.3:1 and 4.9:1 respectively.

Two examples illustrate the pattern:

\begin{figure}[h]
\small
\hrule\vspace{3pt}
\textbf{Emoji (SNLI):}\\
P: a \texttt{[UNK]} selling donuts to a customer during a world exhibition\\
H: a \texttt{[UNK]} selling donuts to a customer.\\
True: \emph{entailment} \quad Baseline: \emph{neutral} \quad Hybrid: \emph{entailment}\\[5pt]
\textbf{Noise (SNLI):}\\
P: two young boys of opposing teams play football, wearing protection uniforms\\
H: boys play football \textbf{deadass}\\
True: \emph{entailment} \quad Baseline: \emph{contradiction} \quad Hybrid: \emph{entailment}
\vspace{3pt}\hrule
\caption{Two cases where the baseline fails and hybrid recovers. In the
  emoji example, ``[UNK]'' is the actual tokenizer output replacing a
  content word. In the noise example, the baseline treats \emph{deadass}
  as a content modifier.}
\label{fig:examples}
\end{figure}

The noise example is particularly telling. The baseline predicts
\emph{contradiction}, suggesting that it assigns inferential weight to the
appended token rather than treating it as irrelevant. The augmented model
has seen enough training examples of the form ``sentence + \emph{deadass}''
with unaffected labels that it has learned the token carries no inferential
weight.

\section{Discussion}

\subsection{What the Failure Mode Tells You About the Fix}

The clearest result in the mitigation tables is not the top-line numbers
but the ordering: preprocessing beats augmentation on emoji, augmentation
beats preprocessing on noise. That pattern holds across both models and
both datasets, though the preprocessing advantage on ELECTRA MultiNLI
emoji is marginal (81.50\% vs.\ 81.48\%).

The explanation follows from where each transform does damage. Emoji
removes lexical information at the tokenizer, by the time the encoder
runs, those tokens are \texttt{[UNK]} and there is nothing for
augmentation to teach the model about. Preprocessing removes the problem
before it reaches the tokenizer. Noise is the opposite: the tokens are
in-vocabulary and fully visible to the encoder, but the model has no
training context for them in NLI. A filter list can strip known noise
tokens, but it will not generalize to new ones. Augmentation builds
invariance through exposure rather than enumeration.

\subsection{Why Hybrid Works as a Default}

Hybrid applies preprocessing before tokenization and trains on augmented
data, so it addresses both failure modes simultaneously. In practice, the
two interventions do not interfere with each other: preprocessing does not
make augmented training less effective, and augmented training does not make
preprocessing unnecessary. The empirical result bears this out: for ELECTRA, hybrid is consistently
best across all ten conditions. For RoBERTa on MultiNLI, targeted individual
mitigations can marginally edge hybrid on specific transforms, but hybrid
remains the safest single choice across all conditions simultaneously.

The practical cost is low. Preprocessing adds no training cost and
negligible inference latency. Augmented training adds roughly 50\% more
steps but no new hyperparameters beyond the warmup rate adjustment
needed for larger models. Accuracy on clean text does not drop
($p > 0.05$ in all conditions).

For RoBERTa on MultiNLI,
the picture is more mixed, preprocessing alone marginally edges hybrid
on emoji (90.31\% vs.\ 90.14\%), and augmentation alone marginally edges
hybrid on noise (90.37\% vs.\ 90.17\%). At that scale, the individual
methods are close enough to hybrid that either is defensible for a
specific failure mode. Hybrid is still the safer default if you are not
sure which failure mode dominates your deployment data.

\subsection{The LLM Comparison}

The LLM comparison is worth examining carefully because the direction of
the result is not obvious in advance. One might expect a 14M-parameter
model to be uniformly weaker than GPT-4o-mini; instead, the baseline
ELECTRA is actually \emph{worse} than GPT-4o-mini on transformed variants,
while being better on clean text. This makes sense once you consider that
GPT-4o-mini was likely trained on data containing emoji and informal
language, which is consistent with its implicit robustness that a model
trained only on formal NLI data lacks.

With hybrid training, ELECTRA gains robustness through targeted exposure
and normalization rather than pretraining scale. It outperforms GPT-4o-mini
on SNLI across all variants and matches it on MultiNLI, a model 25$\times$
smaller reaching parity on robustness once given the right training signal.

The gap for the unmitigated baseline is worth noting. It is not a small
degradation on unusual inputs; it is a statistically significant drop on
the kinds of surface forms that appear in real user text. The mitigation
cost is low enough that skipping it is hard to justify.

\subsection{Training Stability at Scale}
\label{sec:stability}

\begin{figure*}[t]
\centering
\includegraphics[width=0.92\linewidth]{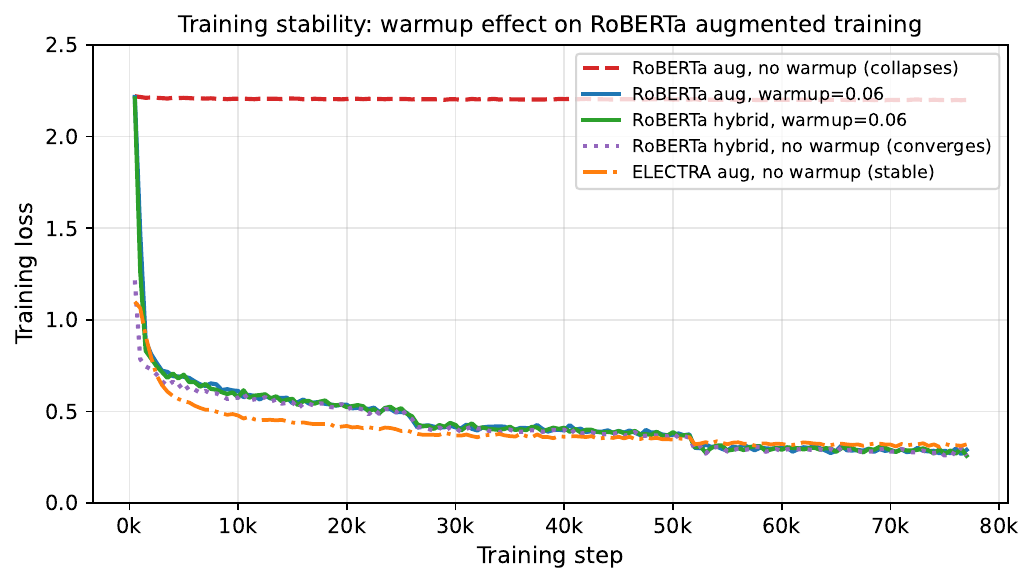}
\caption{Training loss over steps for five RoBERTa-large and ELECTRA-small
  runs. RoBERTa augmented without warmup collapses (flat loss $\approx$2.2,
  accuracy 33.82\%). Adding \texttt{warmup\_ratio=0.06} restores normal
  convergence. RoBERTa hybrid without warmup also converges normally
  (93.00\%), because preprocessing reduces the OOD shock at step~0.
  ELECTRA is stable without warmup in all conditions.}
\label{fig:stability}
\end{figure*}

When we first ran RoBERTa-large augmented training without a warmup
schedule, the model collapsed to 33.82\% accuracy, equivalent to random
guessing for a three-class problem. ELECTRA-small trained without warmup
produced no such collapse. Figure~\ref{fig:stability} shows training
loss curves for five representative runs. The collapsed run maintains a flat
loss of $\approx$2.2 across all 77K steps, while the warmup run drops
normally from 2.2 to 0.30. We believe the combination of a larger model,
a 50\% larger training set, and out-of-distribution examples appearing from
the very first batch led to gradient instability before the model's
representations had time to stabilize. Adding \texttt{warmup\_ratio=0.06}
resolved the issue entirely. Baseline and preprocessing runs, which train on
clean data only, were unaffected.

One additional observation: the hybrid run \emph{without} warmup also
converges normally (final accuracy 93.00\%). The preprocessing step in
hybrid normalizes informal tokens before they reach the model, reducing the
OOD shock in early batches. Augmentation-only exposes the model to raw
informal text from step one, which is precisely the condition that triggers
instability at this scale.

The practical lesson: robustness training with augmented data at large
model scale requires a warmup schedule that standard fine-tuning recipes
often omit.

\subsection{Limitations}

The emoji dictionary and noise token list are finite. Real informal text
uses emoji more creatively than simple noun substitution, a string of
three emoji in sequence conveys something our mapping does not handle —
and conversational tokens evolve faster than any fixed list. Applying
transforms at probability 1.0 is also more extreme than real usage; the
accuracy drops reported here are probably an upper bound on what deployed
models would see. Emoji accuracy drops may also be slightly overestimated:
imperfect recovery (73\% exact) means some examples have a mismatched
label after preprocessing, introducing a small amount of residual label noise
into the evaluation.

The LLM evaluation is zero-shot only. We did not test few-shot prompting
or chain-of-thought, either of which could shift the comparison in either
direction.

The experiments are in English. How tokenization failure rates interact
with non-Latin scripts or morphologically rich languages is a separate
question; character-level tokenizers sidestep the emoji-UNK problem
entirely, and aggressive BPE fragmentation behaves differently across
vocabulary sizes.

\section{Conclusion}

The accuracy drops from emoji and noise look similar on the surface
but come from different places in the pipeline. Emoji triggers
tokenization failure: 93.6\% of emoji examples contain at least one
\texttt{[UNK]} in ELECTRA's WordPiece tokenizer, and even RoBERTa's
BPE tokenizer, which produces no \texttt{[UNK]}, shows 16.7\%
higher subword fragmentation. Noise tokens pass through the tokenizer
intact but carry no meaning in NLI context, leading the model to
treat filler words as content. The confusion matrices reflect this:
emoji errors collapse toward neutral, noise errors drift toward
contradiction.

Preprocessing and augmentation each address one of these failure
modes. Used together, they recover most of the accuracy loss with
no degradation on clean text. Against GPT-4o-mini zero-shot, an
unmitigated fine-tuned model loses badly on transformed inputs
despite being stronger on clean text. Hybrid training reverses that:
a 14M-parameter model reaches parity with a frontier LLM on MultiNLI
and surpasses it on SNLI. The robustness gap is a training choice,
not a capacity constraint.

\bibliographystyle{plainnat}

\onecolumn
\appendix
\section{Transform Examples}
\label{sec:transform-examples}

Table~\ref{tab:transform-examples} shows concrete before-and-after examples
for all four transform types. Examples are drawn from SNLI-style premise--hypothesis
pairs to illustrate the effect of each transform on real NLI inputs.

\begin{table}[H]
\centering
\small
\begin{tabular}{p{0.35\textwidth}lp{0.35\textwidth}}
\toprule
\textbf{Original} & \textbf{Type} & \textbf{Transformed} \\
\midrule
A man is running in the park. & Emoji &
A \textit{[man]} is \textit{[runner]} in the \textit{[park]}. \\
A woman holds a picture of her friend. & Emoji &
A \textit{[woman]} holds a \textit{[photo]} of her \textit{[person]}. \\
\midrule
I am going to call my friend tomorrow. & Slang &
I am gonna call my homie tomorrow. \\
She is trying to take a picture of the dog. & Slang &
She is tryna take a pic of the dog. \\
\midrule
Boys play football. & Noise &
Boys play football \textbf{deadass}. \\
Someone is near the water. & Noise &
Someone is near the water \textbf{no cap}. \\
\midrule
A woman is going to take a picture of her friend. & Combined &
A \textit{[woman]} is gonna take a pic of her homie \textbf{tbh}. \\
Two dogs are running outside. & Combined &
Two \textit{[dog]} are \textit{[runner]} outside \textbf{lowkey}. \\
\bottomrule
\end{tabular}
\caption{Before-and-after examples for each transform type. Emoji substitutions
  shown as bracketed labels. Noise tokens in bold. All examples preserve the gold label.}
\label{tab:transform-examples}
\end{table}

\paragraph{Slang mapping policy.} Phrase-level substitutions are applied
via a deterministic lookup table. Representative entries include:
\emph{going to} $\to$ \emph{gonna},
\emph{trying to} $\to$ \emph{tryna},
\emph{picture} $\to$ \emph{pic},
\emph{friend} $\to$ \emph{homie},
\emph{kind of} $\to$ \emph{kinda},
\emph{want to} $\to$ \emph{wanna}.
All substitutions are meaning-equivalent by construction.

\paragraph{Noise token list.} The noise transform appends one token drawn
uniformly from the following set:
\emph{deadass}, \emph{lowkey}, \emph{no cap}, \emph{tbh}, \emph{highkey},
\emph{on god}, \emph{frfr}, \emph{real talk}, \emph{bet}.
None of these tokens appear in the SNLI or MultiNLI training data.
Table~\ref{tab:noise-glossary} gives the approximate meaning of each token
for readers unfamiliar with Gen-Z slang.

\begin{table}[H]
\centering
\small
\begin{tabular}{ll}
\toprule
\textbf{Token} & \textbf{Approximate meaning} \\
\midrule
deadass   & seriously, I mean it \\
no cap    & no lie, for real \\
tbh       & to be honest \\
lowkey    & kind of, somewhat \\
highkey   & very much, definitely \\
frfr      & for real for real (emphasis) \\
on god    & I swear, I'm serious \\
real talk & honestly speaking \\
bet       & agreed, understood \\
\bottomrule
\end{tabular}
\caption{Gen-Z filler tokens used in the noise transform and their
approximate meanings. All are sentiment-neutral and carry no inferential
weight relevant to NLI.}
\label{tab:noise-glossary}
\end{table}

\end{document}